\documentclass[sigconf]{acmart}
\usepackage{popets}

\setcopyright{popets}
\copyrightyear{YYYY}

\acmYear{YYYY}
\acmVolume{YYYY}
\acmNumber{X}
\acmDOI{XXXXXXX.XXXXXXX}
\acmISBN{}
\acmConference{Proceedings on Privacy Enhancing Technologies}
\usepackage{graphicx}
\usepackage{placeins}
\usepackage{xcolor}
\usepackage{tabularx}
\usepackage{arydshln}
\usepackage{etoolbox}
\usepackage{xspace}
\usepackage{pifont}
\usepackage{soul}
\usepackage{makecell}

\newcommand{\Xin}{\mathbf{X}}
\newcommand{\Yout}{\mathbf{Y}}
\newcommand{\Ypriv}{\mathbf{Z}}
\newcommand{\Zpred}{\widehat{\Ypriv}}
\newcommand{\Ypred}{\widehat{\Yout}}
\newcommand{\Yattack}{\Yout^{\mathrm{rec}}}

\newcommand{\encoder}{\mathcal{E}}
\newcommand{\decoder}{\mathcal{D}}
\newcommand{\name}{\textup{\textsc{Decoy}}\xspace}
\newcommand{\circled}[1]{\textcircled{\scriptsize #1}}

\newtheorem{assumption}{Assumption}

\newif\ifshownotes
\shownotesfalse

\ifshownotes
  \newcommand{\mynote}[3]{\textcolor{#2}{\textsf{\small[\textbf{#1}: #3]}}}
  \AtBeginDocument{\typeout{^^J%
***************************************************^^J%
* AUTHOR NOTES ARE VISIBLE (\string\shownotestrue) *^^J%
***************************************************^^J}}
\else
  \newcommand{\mynote}[3]{}
\fi

\newcommand{\lc}[1]{\mynote{Lydia}{magenta}{#1}}
\newcommand{\gs}[1]{\mynote{Giulio}{red}{#1}}

\newcommand{\ab}[1]{\mynote{Andrea}{blue}{#1}}

\begin{document}

\title{\name{}: Preserving Predictive Utility While Limiting Adversarial Reconstruction}


\author{Giulio Segalini}
\affiliation{%
  \institution{Université de Neuchâtel}
  \city{Neuchâtel}
  \country{Switzerland}
}
\email{giulio.segalini@unine.ch}

\author{Zhi Wen Soi}
\affiliation{%
  \institution{TU Delft}
  \city{Delft}
  \country{Netherlands}
}
\email{z.w.soi@tudelft.nl}

\author{Aditya Shankar}
\affiliation{%
  \institution{TU Delft}
  \city{Delft}
  \country{Netherlands}
}
\email{a.shankar@tudelft.nl}

\author{Andrea Bertazzi}
\affiliation{%
  \institution{ASM}
  \city{Almere}
  \country{Netherlands}
}
\email{andrea.bertazzi@asm.com}

\author{Lydia Y. Chen}
\affiliation{%
  \institution{Université de Neuchâtel \& TU Delft}
  \city{Neuchâtel}
  \country{Switzerland}
}
\email{lydiaychen@ieee.org}

\begin{abstract}
Data owners conveniently outsource their machine learning tasks to third-party providers, risking the exposure of sensitive data.
A common solution is to encode data to preserve useful patterns while concealing the original values.
However, the same patterns that are preserved for learning may still allow the provider to estimate the original values accurately.
We mitigate this reconstruction risk through the following contributions:
\textit{(i)} we demonstrate the fragility of existing encodings through attacks that reconstruct confidential targets using only their value range;
\textit{(ii)} we propose \name{}, a framework enabling data owners to transform data with worst-case reconstruction-error guarantees under ideal Gaussian encoding, even when the adversary knows the full target distribution;
\textit{(iii)} we quantify the adversary's reconstruction capabilities and derive worst-case reconstruction-error lower bounds under \name{}. 
The core of \name{} is to reshape and secretly rotate the data so that it is uniformly distributed on a high-dimensional sphere, obscuring the original geometry while preserving the patterns required for machine learning.
Across four time-series datasets, attacks on existing privacy methods achieve up to 95\% lower reconstruction error than simply guessing the midpoint of the known target range, while our strongest \name{} variants prevent any improvement over this guess.
Models trained on \name{}'s encoded targets can match or exceed the accuracy of models trained on the original, unprotected targets, or incur a moderate cost.

\end{abstract}


\maketitle

\section{Introduction}
\label{sec:introduction}



A data owner may want to outsource model training to a provider with greater computing resources and modeling expertise while keeping its prediction targets confidential~\cite{melissourgos2021outsourcing,xu2023privateinspection}.
For example, a building operator may predict confidential electricity consumption from public weather and calendar data~\cite{eibl2018loadprivacy}.
A hospital may predict peak demand for cardiovascular care from public air quality measurements while keeping its admission records confidential~\cite{qiu2020peakdemand}.
A financial institution may predict the performance of a confidential portfolio from public market indicators~\cite{liu2025pub2priv}.
The provider needs to learn the relationship between these shared inputs and confidential targets.
One approach is for the data owner to encode the targets locally and keep the encoder and decoder private~\cite{silofuse}.
The provider learns to predict the encoded targets, and the data owner decodes its predictions locally.
This requires an encoding that preserves useful patterns for learning while concealing the original target values.

Existing approaches impose different costs and limitations.
Cryptographic methods protect data during training but require additional computation and communication~\cite{stv,hardy2017private}.
Adding noise to targets can reduce prediction accuracy while leaving patterns that allow the provider to estimate the original values~\cite{labeldp,regressiondp,labeldp_lia}.
Another approach keeps the encoder private and shares only encoded data for learning~\cite{silofuse,neuracrypt}.

\begin{figure}[t]
    \centering
    \includegraphics[width=\linewidth]{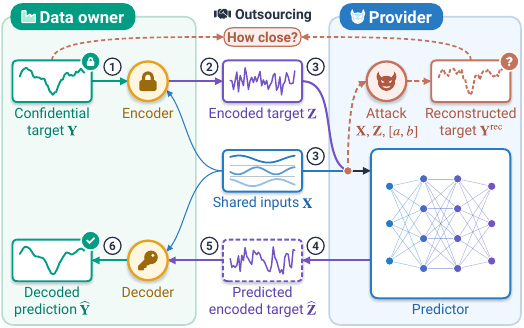}
    \caption{Outsourcing with confidential targets.
    The data owner encodes confidential targets and sends them with the shared inputs; the provider trains a predictor and returns predicted encoded targets, which the data owner decodes.
    Acting as the adversary, the provider also tries to reconstruct the confidential targets.}
    \label{fig:outsourcing}
\end{figure}

Keeping the encoder private, however, can still leave useful clues in the encoded targets.
The same statistical patterns that support learning may allow the provider to estimate confidential values accurately, even without knowing how they were encoded.
Several different original values may be consistent with the encoded data, yet close enough to one another to permit accurate estimation.
In this work, we develop two attacks \lc{introduce the specific terms }\gs{done} that show how a provider can accurately reconstruct confidential targets even when the encoder is kept private.
The range-only PCA attack reads target levels off the dominant direction of the encoded targets, and the predictor-denoising attack averages out added noise through prediction.
We study a provider that knows the type of encoding and a valid range for the target values, but has no original target examples or access to the data owner's encoder.
The challenge of outsourcing ML training is to design encodings that limit reconstruction accuracy while preserving predictive utility.


We introduce \name{}, a framework that reshapes confidential targets before applying a secret rotation.
Reshaping makes the encoded distribution look similar in every direction, making the rotation harder to identify. \lc{Add one or two sentences to summarize the pipeline of Decoy. }\gs{done}
\name{} centers the targets, reshapes them toward a standard Gaussian distribution, and whitens them, then applies the secret rotation and removes each vector's radius.
The encoded targets are approximately uniformly distributed on the unit sphere, and the data owner reverses these stages to decode the provider's predictions.
Under ideal Gaussian encoding assumptions, we derive a lower bound on the provider's reconstruction error, even when it knows the target distribution.
The proof rests on the provider's uncertainty about which original targets produced the encoded data.
We evaluate the privacy and prediction accuracy of practical encoders empirically.


We evaluate \name{} against five existing defenses on four time-series datasets using two neural predictors.
The provider predicts a confidential target channel from shared input channels over the same time window.
We measure both the provider's accuracy in reconstructing confidential targets and the data owner's accuracy after decoding the provider's predictions.
Across all four datasets, no tested range-only attack on \name{} improves on always predicting the midpoint of the known target range.
In summary, we make the following \textbf{contributions}: 
\begin{itemize}
    \item \textbf{Empirical attacks on outsourcing methods.} 
    We demonstrate that a provider can reconstruct confidential targets from encoded targets using knowledge of their value range.
    Our PCA and predictor-denoising attacks exploit preserved variance directions and input correlations.
    These attacks drastically improve the adversary's reconstructions, leading even to exact recovery.
    \item \textbf{\name{}.} We hide targets by reshaping toward rotational symmetry and removing all radius cues from the encoded targets.
    We use five stages: centering, Gaussian reshaping, whitening, secret rotation, and radius removal, addressing the drawbacks of learned encoders that struggle to balance minimizing reconstruction-errors and output distribution shapes.
    \item \textbf{Theoretical limits on reconstruction.}
    We quantify how accurately an adversary can recover confidential targets from encoded data.
    Our analysis identifies conditions under which the adversary cannot guarantee more accurate reconstruction than always guessing the target mean, even with knowledge of the target distribution.
    \item \textbf{Privacy and prediction accuracy.}
    We evaluate \name{} against five existing defenses on four multivariate time-series datasets using two neural predictors.
    Attacks on existing defenses achieve up to 95\% lower reconstruction mean squared error on average than always predicting the midpoint of the known target range.
    Statistical \name{} and \name{}+AE prevent any tested range-only attack from improving on this baseline while preserving useful predictive information.
\end{itemize}

\section{Related Work}
\label{sec:related-work}

\begin{table}[t]
\centering
\definecolor{supportgreen}{RGB}{25,113,62}
\definecolor{partialamber}{RGB}{150,100,20}
\definecolor{unsupportedred}{RGB}{180,35,35}
\newcommand{\yes}{\textcolor{supportgreen}{\ding{51}}}
\newcommand{\na}{\textcolor{unsupportedred}{\ding{55}}}
\newcommand{\partialmark}{\textcolor{partialamber}{\ding{115}}}
\caption{Learned and parametric encodings: predictor independence, adversary knowledge, and reconstruction analysis in each method's original setting.}
\label{tab:privacy-methods}
\small
\begin{tabular}{lccc}
\toprule
\textbf{Method} & \textbf{\makecell{Predictor\\agnostic}} & \textbf{\makecell{Known\\domain}} & \textbf{\makecell{Recon. bound\\\& evaluation}}\\

\midrule

\multicolumn{4}{l}{\textit{Learned encodings}}\\
\quad SiloFuse~\cite{silofuse} & \yes & \partialmark & \na\\
\quad S-VAE$^{*}$~\cite{svae} & \yes & \na & \na\\
\quad Inf\textsuperscript{2}Guard$^{\dagger}$~\cite{noorbakhsh2024inf2guard} & \partialmark & \yes & \yes\\
\quad Pub2Priv~\cite{liu2025pub2priv} & \na & \yes & \na\\
\quad FedTDD~\cite{fedtdd} & \na & \na & \na\\

\midrule

\multicolumn{4}{l}{\textit{Parametric encodings}}\\
\quad Label-DP~\cite{regressiondp} & \yes & \yes & \na\\
\quad MARVELL~\cite{marvell} & \partialmark & \yes & \na\\
\quad Random proj.~\cite{randproj} & \partialmark & \partialmark & \partialmark\\
\quad Geometric pert.~\cite{geoperturb} & \partialmark & \partialmark & \partialmark\\
\quad Matrix masking~\cite{melissourgos2021outsourcing} & \partialmark & \na & \na\\
\quad InstaHide$^{\dagger}$~\cite{instahide} & \partialmark & \partialmark & \partialmark\\
\quad NeuraCrypt$^{\dagger}$~\cite{neuracrypt} & \partialmark & \yes & \partialmark\\

\midrule

\textbf{\textsc{Decoy}$^{\ddagger}$ (ours)} & \yes & \yes & \yes\\

\bottomrule
\end{tabular}

\vspace{4pt}
\parbox{\columnwidth}{\footnotesize \yes supported; \partialmark partial; \na not covered. In the last column, \yes denotes both a reconstruction-error bound and attack evaluation; \partialmark denotes one of the two. $^{*}$Representation learning without a privacy analysis; $^{\dagger}$Input classification; $^{\ddagger}$Encoded-target regression.}
\end{table}

We summarize prior work according to how the data owner encodes confidential targets before outsourcing, using either learned or parametric encodings, and then review reconstruction attacks against such encodings.
Table~\ref{tab:privacy-methods} compares representative methods from both families.

\noindent\textbf{Learned encodings.}
A learned encoding trains a neural network on the confidential data and releases its outputs.
SiloFuse~\cite{silofuse} keeps each data owner's autoencoder private and shares only its latent representations, and hyperspherical variational autoencoders~\cite{svae} map data onto the unit sphere with sampling noise.
Inf\textsuperscript{2}Guard~\cite{noorbakhsh2024inf2guard} trains representations against reconstruction, membership, and property inference using information-theoretic objectives.
Generative approaches release synthetic time series instead, trained under differential privacy and conditioned on public information~\cite{liu2025pub2priv}, or distilled across data owners with misaligned features and time steps~\cite{fedtdd}.
Because one network must both preserve the learning signal and enforce privacy, each objective limits the other, and the encoded outputs can retain variance directions that track the confidential targets.
Their privacy arguments assume a secret encoder and an unknown data distribution, protect inputs for classification, or bound the influence of each record through differential privacy~\cite{silofuse,noorbakhsh2024inf2guard}.
\emph{These learned encodings thus lack a lower bound on the error of reconstructing confidential targets when the provider knows the data domain.}
 
\noindent\textbf{Parametric encodings.}
A parametric encoding applies a fixed transformation, such as added noise or multiplication by a matrix, whose parameters are sampled at random or set in closed form and kept secret.
Cryptographic protocols are the strongest form: secret sharing masks each value with random shares~\cite{stv}, and homomorphic encryption hides it under a secret key~\cite{hardy2017private}.
They hide the target values entirely, but their communication and computation overheads~\cite{zhou2023bicoptor} make them impractical for outsourced training, and protocols such as STV~\cite{stv} support only specific model families.
Lighter encodings keep the data learnable without decryption.
Additive perturbation adds zero-mean noise to each value~\cite{agrawal2000privacy}, and label differential privacy randomizes the targets according to a privacy budget~\cite{labeldp,regressiondp}.
MARVELL~\cite{marvell} similarly adds optimized noise to the gradients exchanged in split learning.
Multiplicative encodings combine the data with secret matrices, as in random projection~\cite{randproj}, geometric perturbation with rotation, translation, and noise~\cite{geoperturb}, and matrix masking that preserves feature means and correlations~\cite{melissourgos2021outsourcing}.
Image encodings follow the same pattern, mixing images with random sign flips in InstaHide~\cite{instahide} or passing them through a random, untrained network in NeuraCrypt~\cite{neuracrypt}.
Each transformation must keep the signal that the provider's predictor learns.
Zero-mean noise averages out during training, so the predictor still recovers the part of each target explained by the shared inputs.
Likewise, Wu et al.~\cite{labeldp_lia} show that input--target relationships preserved under label differential privacy still support accurate label inference.
Secret matrices preserve distances, inner products, or dominant variance directions, which remain visible to the provider without the key.
\emph{They protect privacy by adding noise, setting a privacy budget, or keeping their parameters secret, but do not bound the worst-case reconstruction error.}
 
\noindent\textbf{Reconstruction attacks and bounds.}
Existing attacks exploit the structure that encodings preserve, but assume knowledge beyond the data domain.
Against secret orthogonal transformations, Liu et al.~\cite{liu2006distanceattack} apply PCA and align the principal directions with plaintext reference samples, and Giannella et al.~\cite{giannella2013fewinputs} reconstruct records from a few known inputs.
Wasserstein--Procrustes~\cite{wassersteinprocrustes} aligns rotated data with reference data without known pairings.
Chen and Liu~\cite{geoperturb} rescale ICA reconstructions using known minima and maxima, but match the recovered components to attributes through their known distributions.
Sun et al.~\cite{sun2022forward} use the leading singular direction of forward embeddings to infer binary labels in vertical federated learning.
Attacks on InstaHide and NeuraCrypt combine repeated encodings or match encodings to known original images~\cite{instahideattack,neuracryptattack}, and gradient inversion attacks recover training data from exchanged gradients or activations~\cite{zhu2019dlg,zhao2020idlg,erdogan2022unsplit,yang2025uifv,huang2025gidm,segalini2026goliath}.
These attacks require reference samples, known pairs, known attribute distributions, or binary labels, and do not reconstruct continuous targets from the data domain alone.
Existing bounds on reconstruction error require noise in the encoding or training.
Maeng et al.~\cite{encodinginversion} bound reconstruction error via Fisher information for randomized encoders.
Balle et al.~\cite{reconinformed} bound reconstruction by informed adversaries under differential privacy, and Xiao et al.~\cite{learnableenc} analyze learnable encodings that combine masking with noise.
Guerra-Balboa et al.~\cite{guerrabalboa2026disclosure} bound reconstruction advantage under differential privacy, accounting for the adversary's auxiliary knowledge.
Their measure isolates the increase in reconstruction success caused by a record's participation.

\emph{\name{} is the only method in Table~\ref{tab:privacy-methods} that is predictor-agnostic, covers a provider that knows the data domain, and provides both a reconstruction-error bound and an attack evaluation.
Our analysis bounds worst-case reconstruction error over compatible target sets and private encoders under ideal Gaussian encoding, without adding noise.}

\begin{figure*}[t]
    \centering
    \includegraphics[width=\textwidth]{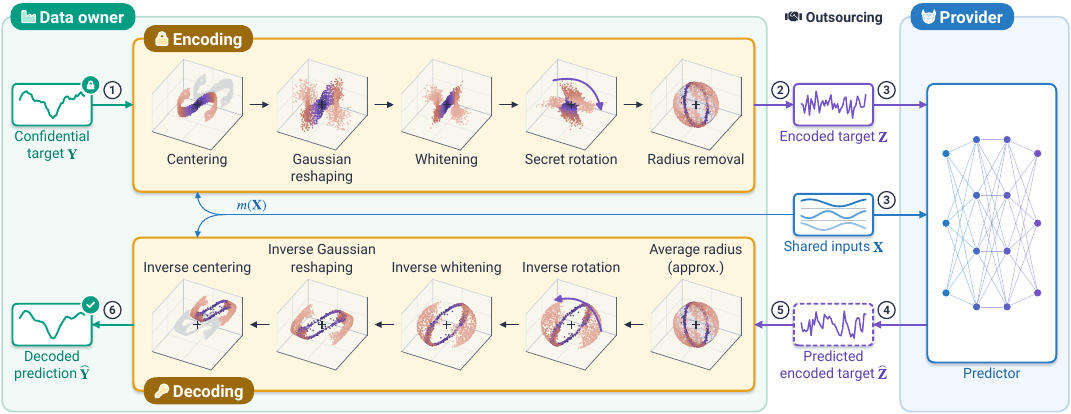}
    \caption{\name{} encoding, outsourced training, and decoding pipeline.
    The data owner transforms confidential targets $\Yout$ into encoded targets $\Ypriv$; the provider learns to predict $\Ypriv$ from input features $\Xin$.
    Data owner converts the predicted encoded targets $\Zpred$ returned by the provider into decoded predictions $\Ypred$.} 
    \label{fig:problem}
\end{figure*}

\section{Risk of Leaking Confidential Targets} 
\label{sec:attacks-overview}
Existing methods transform confidential targets to hide their values while keeping them useful for learning.
This can be achieved through learned methods, such as autoencoders~\cite{silofuse}, or parametric methods, such as random projection~\cite{randproj}, geometric perturbation~\cite{geoperturb}, and additive noise~\cite{agrawal2000privacy}.
However, the transformed data can still reveal the original values, when combining the range information.

Under this assumption, we design two reconstruction attacks.
The \textbf{range-only PCA attack} targets learned and linear encodings, e.g., autoencoders and random projection, and the \textbf{predictor-denoising attack} targets additive noise.

\subsection{Outsourcing Setting and Threat Model}
\label{sec:threat-model}

We consider a single data owner who tries to outsource the training of a regression model to a service provider, shown in Figure~\ref{fig:problem}.
Training requires an input vector $\Xin$ and a target vector $\Yout$.
The target vector is considered confidential and must remain secret in its plain form.
For example, a building operator may outsource forecasting of its daily electricity load, $\Yout$, from the input of public weather and calendar data $\Xin$.

We represent $\Yout\in\mathbb R^T$ as a vector of $T$ real-valued coordinates.
We assume the service provider knows a range containing $\Yout$, though exact values are kept secret.
Formally, $\Yout\in[a,b]^T$, where $a<b$ are valid bounds for every coordinate, such as sensor limits or the rated capacity of a grid connection.
The provider's known range may be wider, $[a-\Delta,b+\Delta]$, where $\Delta\geq 0$.

The data owner keeps a private encoder $\encoder$, which can be learned or parametric, and a decoder $\decoder$, which reverses the transformation.
Before sending the training data to the provider, it encodes each target as $\Ypriv_i=\encoder(\Yout_i)$, where $i$ indexes the $n$ training records.
It collects the training pairs into a set $\mathcal O=\{(\Xin_i,\Ypriv_i)\}_{i=1}^n$.
These are sent to the provider, which learns to predict $\Ypriv$ from $\Xin$.
At inference time, a new input $\Xin'$ is sent to the provider, which then returns a predicted encoded target $\Zpred$.
The data owner then decodes it locally to obtain the prediction $\Ypred=\decoder(\Zpred)$.

We assume an honest-but-curious provider that follows the training protocol but also attempts to reconstruct the confidential targets from the information it receives.
The provider knows the encoding scheme, but the encoder's parameters remain private to the data owner.
From $\mathcal O$ and the known range, it tries to reconstruct the confidential $\Yout_i$.

\subsection{Reconstruction from Encoded Targets}
\label{sec:attacks}
\begin{figure}[t]
\centering
\includegraphics[width=\columnwidth]{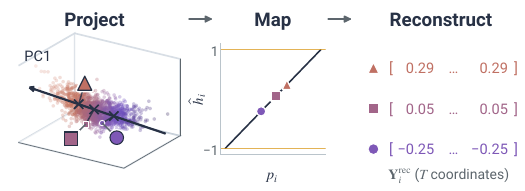}
\input{figures/generated/pca_attack_caption}
\label{fig:pca-attack}
\end{figure}

Encoding can hide individual target values while preserving differences between records.
For example, an encoding of daily electricity load must still reflect higher electricity consumption on weekdays than on holidays.
Preserving such patterns enables learning, but also risks privacy exposure, even when the encoder is secret.
Our first attack exploits this to estimate the overall level of each confidential target.

We define the \emph{level} of a target as the mean of its coordinates, $h(\Yout_i)=\frac{1}{T}\sum_{t=1}^{T}Y_{i,t}$.
Our \textbf{range-only PCA attack} estimates this level using only the encoded targets and the known range, as shown in Figure~\ref{fig:pca-attack}.
The intuition is that, when differences in target levels dominate the variation across encoded records, this variation reveals their relative levels.
To exploit this, the provider applies principal component analysis (PCA) to the encoded targets and projects each onto the direction of greatest variation, obtaining a score $p_i$.
Suppose the target levels span $[a,b]$, while the provider knows only the range $[a-\Delta,b+\Delta]$.
The provider linearly maps the PCA scores to this range:
\[
\widehat h_i=a-\Delta+(b-a+2\Delta)\frac{p_i-p_{\min}}{p_{\max}-p_{\min}},
\]
where $p_{\min}$ and $p_{\max}$ are the smallest and largest scores (if they coincide, $\widehat h_i=(a+b)/2$).
Finally, the provider reconstructs each target as the constant vector $\Yattack_i=(\widehat h_i,\ldots,\widehat h_i)\in\mathbb R^T$.

The attack requires no knowledge of the encoder's parameters.
However, the PCA direction has an arbitrary sign, so the ordering of the scores may reverse the ordering of the target levels.
Suppose some shift and scaling of the PCA scores approximates every level within $e\ge0$.
Then, after possibly reflecting all estimates around $(a+b)/2$ to correct the arbitrary sign,
\[
|\widehat h_i-h(\Yout_i)|\le 2e+\Delta\quad\text{for every record } i.
\]
The extreme scores lie within $e$ of $a$ and $b$, so mapping them to $[a-\Delta,b+\Delta]$ moves each score by at most $e+\Delta$, on top of its initial error $e$.
For an exact linear relationship, $e=0$, so the error is at most $\Delta$, and the attack recovers the levels exactly when $\Delta=0$.
Appendix~\ref{sec:pca-bound} gives a condition on the encoded targets under which PCA approximates the levels and bounds the resulting error for the full targets.
Although nonlinear encoders may distort this relationship, the attack is effective against some (see Section~\ref{sec:pitfalls}).

\subsection{Reconstruction through Model Prediction}
\label{sec:predictor-denoising}

While the range-only PCA attack exploits variation across encoded targets, our second attack exploits the information learned by the provider's predictor.
Consider additive noise~\cite{agrawal2000privacy}, where $\Ypriv=\Yout+\eta$ and $\eta\in\mathbb R^T$ is zero-mean noise independent of the shared inputs $\Xin$.
The provider trains a regression model to predict the encoded targets by minimizing squared error.
Under this objective, the optimal predictor is the conditional mean of the encoded targets:
\[
\mathbb E[\Ypriv\mid\Xin]
=\mathbb E[\Yout\mid\Xin]+\mathbb E[\eta\mid\Xin]
=\mathbb E[\Yout\mid\Xin].
\]
Thus, the predictor learns the part of the confidential targets that can be inferred from the shared inputs, while averaging out the added noise.
For example, a model trained to predict noisy electricity consumption from weather data can still learn the expected consumption for each weather condition.

Our \textbf{predictor-denoising attack} uses the provider's predictions on the observed inputs as initial reconstructions, $\Yattack_i=\Zpred_i$, since the regression model predicts target values in their original units under additive noise.
Their accuracy depends on how well the shared inputs predict the confidential targets.
For a record from an independent recording, Appendix~\ref{sec:denoising-bound} gives the condition under which the fitted predictor improves on guessing the midpoint.
The provider also considers the encoded targets $\Ypriv_i$ directly or clipped to $[a-\Delta,b+\Delta]^T$, retaining record-specific variations that shared inputs do not explain.

\subsection{Empirical Reconstruction Risk}
\label{sec:pitfalls}
\begin{figure}[t]
\centering
\includegraphics[width=\columnwidth]{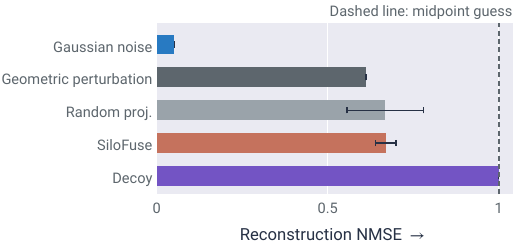}
\caption{Reconstruction NMSE under the best evaluated attack (lower means greater leakage; at $1.0$, dashed, no tested attack beats the midpoint guess).
For each defense, dataset, and seed, we select the most accurate applicable reconstruction among range-only PCA, predictor denoising, direct or clipped encoded targets, and the midpoint guess.
Scores are averaged across FaceDetection, FingerMovements, Heartbeat, and PEMS-SF, weighted by the number of evaluated target values; bars show mean $\pm$ SD over ten seeds.}
\label{fig:attack-teaser}
\end{figure}

We evaluate reconstruction risk on four time-series datasets against SiloFuse~\cite{silofuse}, random projection~\cite{randproj}, geometric perturbation~\cite{geoperturb}, and additive Gaussian noise~\cite{agrawal2000privacy} (Figure~\ref{fig:attack-teaser}).
The range-only PCA attack is applied to SiloFuse, random projection, and geometric perturbation, and the predictor-denoising attack to additive Gaussian noise.
We measure reconstruction accuracy using normalized mean squared error (NMSE): the MSE of the most accurate tested reconstruction, $\frac1n\sum_{i=1}^n\|\Yattack_i-\Yout_i\|_2^2$, divided by the MSE of always predicting the midpoint $(a+b)/2$.
An NMSE of $1$ means no tested attack improves upon this trivial guess, while lower values indicate greater leakage.
To account for the arbitrary sign of the PCA direction, we evaluate reconstructions using both $\widehat h_i$ and its reflection $a+b-\widehat h_i$.
We report the lower reconstruction error, resolving the sign ambiguity in the adversary's favor.

The attacks reveal substantial leakage across existing methods.
SiloFuse's learned nonlinear encoder and a linear random projection both average an NMSE of about $0.67$, showing that nonlinearity does not necessarily conceal differences in target levels.
Geometric perturbation reaches $0.61$, as its secret rotation preserves the variation exploited by the range-only PCA attack.
Additive Gaussian noise is particularly vulnerable, reaching $0.05$ on average and below $0.01$ on some datasets: the provider's predictor can recover the information preserved for learning despite the added noise.

These attacks exploit two forms of preserved information: differences in target levels across records and relationships between shared inputs and confidential targets.
Secret rotation preserves the variation used by PCA, while independent additive noise can be averaged out through prediction.
These failures motivate changing the structure exposed by encoded targets while retaining useful information for learning.

\section{\name{}}

\label{sec:method}

To prevent the attacks of Section~\ref{sec:attacks}, we need an encoder $\encoder$ that conceals the distributional characteristics of the confidential targets $\Yout$.
The encoded targets $\Ypriv$ must not reveal clues about the confidential targets, such as distribution skewness or the directions of the principal components.
\name{} therefore removes shape information and enforces rotational symmetry by mapping the targets onto a standard Gaussian distribution before applying a secret rotation.
Normalizing each rotated vector by its radius then removes the radius, yielding encoded targets that are approximately uniformly distributed on the \textit{unit sphere}~\cite{mardiajupp}.   

\subsection{Framework Overview}
\label{sec:encoding-procedure}
The first three stages of \name{} are fitted to the private training set and bring the targets close to a standard Gaussian distribution.
Given this distribution, privacy rests on the secret rotation and radius removal, which do not depend on the data.

Figure~\ref{fig:problem} shows the complete workflow in six steps.
Before sharing data, the data owner fits the encoding stages on its private training set and samples one secret orthogonal matrix $Q\in O(k)$ shared across records.
For each training record, the data owner passes the confidential target $\Yout$ through the five stages to obtain the encoded target $\Ypriv=\encoder(\Yout)$ (\circled{1}).
It sends $\Ypriv$ to the provider (\circled{2}).
The data owner also shares the corresponding inputs $\Xin$ (\circled{3}), and the provider trains a predictor from $\Xin$ to $\Ypriv$.
At inference time, the provider maps a new shared input $\Xin$ to a predicted encoded target $\Zpred$ (\circled{4}) and returns $\Zpred$ to the data owner (\circled{5}).
The data owner obtains the decoded prediction $\Ypred=\decoder(\Zpred)$ by reversing the five stages from right to left (\circled{6}).
It scales $\Zpred$ by the mean training radius $\bar r$ in place of the removed radius and adds back the local centering prediction $m(\Xin)$.

Since \name{} transforms only the target vectors $\Yout$, it places no assumptions on the modality or structure of the shared inputs $\Xin$ or on the provider's model architecture.
Any regression or generative predictor that maps $\Xin$ to a $k$-dimensional vector can be trained directly on the pairs $\{(\Xin_i,\Ypriv_i)\}_{i=1}^n$ without access to the data owner's decoder $\decoder$.

\subsection{Privatization Design}
\label{sec:defense}

\begin{table}[t]
\centering
\caption{Stages of the \name{} pipeline and the cue each removes.}
\label{tab:pipeline}
\small
\begin{tabularx}{\columnwidth}{@{}ll>{\raggedright\arraybackslash}X@{}}
\toprule
\textbf{Stage} & \textbf{Output} & \textbf{Statistical cue removed} \\
\midrule
1. Centering & $\mathbf e\in\mathbb R^T$ & Linear correlation with $\Xin$ \\
2. Gaussian reshaping & $\mathbf u\in\mathbb R^k$ & Marginal skewness and bounds \\
3. Whitening & $\mathbf v\in\mathbb R^k$ & Cross-coordinate correlation \\
4. Secret rotation & $\mathbf w\in\mathbb R^k$ & Coordinate orientation \\
5. Radius removal & $\Ypriv\in\mathbb S^{k-1}$ & Radius $\|\mathbf w\|_2$ \\
\bottomrule
\end{tabularx}
\end{table}

Table~\ref{tab:pipeline} summarizes the five stages of \name{} and the statistical cue each stage removes.
The order of stages is fixed because each stage establishes the distribution property required by the next.
Centering is first so that distribution shaping acts only on the residual variation in $\Yout$ beyond linear trends in $\Xin$.
Gaussian reshaping and whitening hide any skewness and the principal axes.
Secret rotation then acts on an approximately Gaussian vector $\mathbf v\approx\mathcal N(0,I_k)$, concealing the coordinate orientation while preserving distances between records.
Finally, the result is projected onto the unit sphere by removing the remaining radius cue.

\noindent\textbf{Centering.}
Shared inputs $\Xin$ often explain coarse variation in the confidential targets $\Yout$, such as baseline offsets or linear trends.
If this predictable component remains in $\Yout$, the provider can correlate $\Xin$ with $\Ypriv$ to narrow down the target values.
To remove this cue before encoding, the data owner fits a lightweight centering model $m:\Xin\to\mathbb R^T$ on its training pairs $(\Xin,\Yout)$ and subtracts its prediction from each target:
\[
\mathbf e=\Yout-m(\Xin).
\]
Subsequent stages act on the residual $\mathbf e$, leaving the nonlinear relationship between $\Xin$ and $\mathbf e$ for the provider's predictor.
In our experiments, $m$ is a ridge regression model fitted locally on summary statistics of the shared inputs, and the data owner inverts centering during decoding by adding $m(\Xin)$ back to the decoded residual $\widehat{\mathbf e}$.
We suppress the dependence on the shared input $\Xin$ in the encoder and decoder notation; the fitted centering model is fixed before encoding.

\noindent\textbf{Gaussian reshaping.}
Even after centering, the residual $\mathbf e$ typically has bounded support, skewness, or heavy tails that would remain visible after rotation and allow the provider to align the rotated point cloud with the data domain.
Simple standardization adjusts the mean and standard deviation of each coordinate, leaving these higher-order shape cues intact.
To remove them, the data owner applies an invertible reshaping map $g:\mathbb R^T\to\mathbb R^k$,
\[
\mathbf u=g(\mathbf e),
\]
that transforms each coordinate toward a standard Gaussian distribution while allowing the data owner to map predicted vectors $\widehat{\mathbf u}$ back to residuals $\widehat{\mathbf e}=g^{-1}(\widehat{\mathbf u})$ during decoding.
By default, \name{} implements $g$ as a coordinate-wise quantile transformation ($k=T$), mapping each coordinate $e_t$ through its empirical cumulative distribution function $\widehat F_t$ and the standard normal quantile function $\Phi^{-1}$, $u_t=\Phi^{-1}(\widehat F_t(e_t))$.
This stage can also be instantiated by a learned normalizing flow (\name{}+NF, $k=T$)~\cite{nice,nsf}, a Gaussian-regularized autoencoder (\name{}+AE, $k\le T$)~\cite{kolouri2018sliced}, or a Gaussian variational autoencoder (\name{}+VAE, $k\le T$)~\cite{kingma2013vae}.

\noindent\textbf{Whitening.}
Making each coordinate of $\mathbf u$ marginally Gaussian does not make the joint distribution rotationally symmetric, because correlations across coordinates stretch the point cloud along dominant axes.
If rotated directly, these directions of greatest variance remain identifiable by the PCA attack of Section~\ref{sec:attacks}.
To equalize spread across all directions, the data owner estimates the training mean $\widehat{\boldsymbol\mu}$ and covariance matrix $\widehat{\Sigma}$ of $\mathbf u$ and transforms each vector as
\[
\mathbf v=\widehat{\Sigma}^{-\frac{1}{2}}(\mathbf u-\widehat{\boldsymbol\mu}).
\]
Multiplying by $\widehat{\Sigma}^{-1/2}$ scales each principal direction by the inverse of its standard deviation, producing a representation with zero mean and unit covariance $I_k$.
\name{} estimates $\widehat{\Sigma}$ using Ledoit--Wolf shrinkage~\cite{ledoitwolf} to keep the inverse square root well-conditioned, and reverses the transformation during decoding via $\widehat{\mathbf u}=\widehat{\Sigma}^{1/2}\widehat{\mathbf v}+\widehat{\boldsymbol\mu}$.
In \name{}+NF, \name{}+AE, and \name{}+VAE, the learned encoder already optimizes all $k$ coordinates jointly toward $\mathcal N(0,I_k)$, so separate whitening is omitted ($\mathbf v=\mathbf u$).

\noindent\textbf{Secret rotation.}
Although the data owner keeps all stage parameters private, centering, reshaping, and whitening are fitted to the training data distribution, so a provider that knows the data domain could approximate their effect.
To introduce a secret key independent of the data, the data owner samples one orthogonal matrix $Q\in O(k)$ uniformly at random and applies it to every whitened vector:
\[
\mathbf w=Q\mathbf v,\qquad Q^\top Q=I_k.
\]
Since $\mathbf v$ approximately follows the standard Gaussian distribution $\mathcal N(0,I_k)$, multiplying by $Q$ leaves the distribution of $\mathbf w$ unchanged while hiding the coordinate axes of $\mathbf v$.
Applying the same orthogonal matrix $Q$ across all records preserves pairwise Euclidean distances and inner products ($\|\mathbf w_i-\mathbf w_j\|_2=\|\mathbf v_i-\mathbf v_j\|_2$), keeping the regression geometry intact for training at the provider.
During decoding, the data owner reverses the rotation by multiplying by $Q^\top$.

\noindent\textbf{Radius removal.}
Since every orthogonal transformation preserves Euclidean length ($\|\mathbf w\|_2=\|\mathbf v\|_2$), a rotated vector still reveals its radius, which measures the magnitude of the residual after reshaping and whitening.
To remove this radius cue, the data owner normalizes each nonzero vector to unit length:
\[
\Ypriv=\frac{\mathbf w}{\|\mathbf w\|_2}.
\]
Normalizing $\mathbf w$ onto the unit sphere $\mathbb S^{k-1}$ retains the direction and discards the radius.
In the ideal case $\mathbf w\sim\mathcal N(0,I_k)$, the direction $\Ypriv$ is uniformly distributed on $\mathbb S^{k-1}$ and independent of the radius~\cite{mardiajupp}.
Hence, the data owner cannot invert this step exactly during decoding.
Instead, it scales each predicted encoded target $\Zpred$ by the average radius $\bar r=\frac{1}{n}\sum_{j=1}^n\|\mathbf w_j\|_2$ of the training vectors, $\widehat{\mathbf w}=\bar r\,\Zpred$.
If the direction is predicted exactly, $\bar r$ is the constant radius that minimizes the mean squared error in $\mathbf w$ over the training vectors.
Radius removal is therefore the only stage designed to discard information, and the data owner reverses all other stages during decoding.

\section{Privacy Analysis}
\label{sec:theory}

We derive lower bounds on the error of any reconstruction attack by the provider against \name{}'s two key features: the secret rotation (Section~\ref{sec:ambiguity}) and radius removal (Section~\ref{sec:leakage}).
The proofs construct \emph{indistinguishable target sets}, which differ from the true targets but yield the same encoded data.
The provider cannot tell which of these sets is the real one, because its observations are identical in every case.
The provider's reconstruction is therefore identical for all these sets.
Since these sets assign different values to the same record, this single reconstruction cannot match all of them: for at least one set, its error is at least the spread of these values, which gives the lower bound.
After radius removal, no attack can guarantee a lower error than guessing the average target for every record.

\subsection{Reconstruction Error Before Radius Removal}
\label{sec:ambiguity}

%
Here we consider \name{} before the step of radius removal.
The data owner encodes each target, reshapes the encoding toward a standard Gaussian distribution, and applies a secret rotation before sending it to the provider.
We ask how accurately the provider can reconstruct the targets from these rotated encodings.
To answer this, we assume ideal encoding: the encodings are exactly standard Gaussian, and the decoder recovers the targets exactly.
A standard Gaussian looks the same after any rotation, so the provider cannot tell which rotation the owner used.
Each possible rotation decodes to a different indistinguishable target set (Proposition~\ref{thm:ambiguity}).
Rotation does not change the length of an encoding, however, so all these sets share each record's radius.
The provider's reconstruction error is therefore at least the spread of the targets with the same radius (Theorem~\ref{thm:reconstruction-bound}).

\begin{assumption}[Ideal encoding]
\label{ass:encoding}
Let $\Yout$ follow the target distribution.
The encoder $\encoder:[a,b]^T\to\mathbb R^k$ and decoder $\decoder:\mathbb R^k\to\mathbb R^T$ satisfy:
\begin{enumerate}
\item[(a)] $\decoder(\encoder(\Yout))=\Yout$ with probability one.
\item[(b)] $\encoder(\decoder(\Ypriv))=\Ypriv$ with probability one for $\Ypriv\sim\mathcal N(0,I_k)$.
\item[(c)] $\encoder(\Yout)\sim\mathcal N(0,I_k)$.
\end{enumerate}
\end{assumption}

Conditions (a) and (b) require the encoder and decoder to invert each other, as in a perfect autoencoder~\cite{he2024manifold}.
Condition (c) adds exact Gaussian reshaping.


We now formally define the target sets that the provider cannot distinguish from the true targets, given its observations $\mathcal O$.


\begin{definition}[Indistinguishable target sets]
\label{def:compatible-encodings}
A target set $\{\widetilde{\Yout}_i\}_{i=1}^n$ is \emph{indistinguishable} from the true targets given $\mathcal O$ if:
\begin{enumerate}
\item[(i)] each $\widetilde{\Yout}_i\in[a,b]^T$ follows the original target distribution;
\item[(ii)] some encoder $\widetilde{\encoder}$ and decoder $\widetilde{\decoder}$ satisfying Assumption~\ref{ass:encoding}, used in place of the owner's in \name{}, produce exactly the observed $\Ypriv_i$ from every $\widetilde{\Yout}_i$ with probability one.
\end{enumerate}
\end{definition}
Without radius removal, condition (ii) reads $\widetilde{\encoder}(\widetilde{\Yout}_i)=\Ypriv_i$; after radius removal (Section~\ref{sec:leakage}), it reads $\widetilde{\encoder}(\widetilde{\Yout}_i)/\|\widetilde{\encoder}(\widetilde{\Yout}_i)\|_2=\Ypriv_i$.

\begin{proposition}[Indistinguishability under rotation]
\label{thm:ambiguity}
Suppose Assumption~\ref{ass:encoding} holds and each target $\Yout_i$ follows the target distribution.
For any rotation $Q$, the targets $\widetilde{\Yout}_i=\decoder(Q^\top\Ypriv_i)$ form an indistinguishable target set (Definition~\ref{def:compatible-encodings}).
If $Q\ne I_k$, every one of them differs from its true target with probability one.
\end{proposition}
 
\begin{proof}
Set $\widetilde{\encoder}=Q\encoder$ and $\widetilde{\decoder}=\decoder\circ Q^\top$.
Since $\mathcal N(0,I_k)$ is unchanged by rotation and the transpose undoes the rotation, $(\widetilde{\encoder},\widetilde{\decoder})$ satisfies Assumption~\ref{ass:encoding}.
Exact inversion gives
\[
\widetilde{\encoder}(\widetilde{\Yout}_i)=Q\encoder(\decoder(Q^\top\Ypriv_i))=\Ypriv_i
\]
with probability one, and each $\widetilde{\Yout}_i$ retains the target distribution and lies in $[a,b]^T$.
If $\widetilde{\Yout}_i=\Yout_i$, exact inversion implies that the rotation leaves $\Ypriv_i$ unchanged.
A rotation other than the identity leaves unchanged only the vectors of a subspace of dimension at most $k-1$, which has probability zero under $\mathcal N(0,I_k)$.
Thus all $n$ targets differ from the true ones with probability one.
\end{proof}
 
Under these assumptions, the provider cannot uniquely identify the confidential targets, even when it knows the target distribution. However, the provider can still recover the radius of each encoded target, since orthogonal rotations preserve length. We let
\[
R=\|\encoder(\Yout)\|_2,\qquad r_i=\|\encoder(\Yout_i)\|_2.
\]
As the rotation varies, the rotated encoding of record $i$ ranges over the sphere of radius $r_i$, so its indistinguishable target $\widetilde{\Yout}_i$ ranges over
\[
\{y=\decoder(z) : \|z\|_2=r_i\}.
\]
These targets may still be close together, so indistinguishability alone does not quantify reconstruction error.
We measure their spread by $\operatorname{Var}(\Yout\mid R=r_i)$, where variance sums the coordinate variances.
The following theorem shows that every attack has at least one indistinguishable target set on which its reconstruction MSE is at least the expected conditional target variance.
Here and in the following subsection, reconstruction MSE averages over records, data, and attack randomness.
 
\begin{theorem}[Bound before radius removal]
\label{thm:reconstruction-bound}
Under the conditions of Proposition~\ref{thm:ambiguity}, for any attack producing reconstructions $\Yattack_i$ from $\mathcal O$ with finite MSE, there exists an indistinguishable target set $\widetilde{\Yout}$ such that
\begin{equation}
\label{eq:reconstruction-bound}
\mathrm{MSE}\bigl(\Yattack,\widetilde{\Yout}\bigr)
\ge\mathbb E\bigl[\operatorname{Var}(\Yout\mid R)\bigr].
\end{equation}
\end{theorem}
 
\begin{proof}
Fix $\mathcal O$ and the attack's reconstructions $\Yattack_i$.
Every indistinguishable target set gives the provider these same observations, so the reconstructions stay fixed as the rotation varies.
We average the error over a uniformly sampled $Q\in O(k)$, independent of the data and the attack.
 
For each record, $Q^\top\Ypriv_i$ is uniform on the sphere of radius $r_i$.
Under Assumption~\ref{ass:encoding}, decoding this random point gives the target distribution conditional on $R=r_i$.
The expected squared error therefore splits into the spread of these targets and the squared distance from the reconstruction to their conditional mean:
\[
\begin{aligned}
&\mathbb E\bigl[\|\Yattack_i-\widetilde{\Yout}_i\|_2^2
  \mid\mathcal O,\Yattack_i\bigr]\\
&\quad=\operatorname{Var}(\Yout\mid R=r_i)
  +\|\Yattack_i-\mathbb E[\Yout\mid R=r_i]\|_2^2.
\end{aligned}
\]
The second term is nonnegative, so the average error over rotations is at least $\operatorname{Var}(\Yout\mid R=r_i)$.
Averaging over records, data, and attack randomness gives the lower bound $\mathbb E[\operatorname{Var}(\Yout\mid R)]$.
Since the average over rotations is at least this large, at least one rotation gives an indistinguishable target set satisfying Eq.~\eqref{eq:reconstruction-bound}.
\end{proof}
 
The bound equals the error of estimating each target by the mean of targets with the same encoded radius.
The provider cannot guarantee a lower reconstruction MSE across all indistinguishable target sets.
 
\begin{figure}[t]
\centering
\includegraphics[width=\columnwidth]{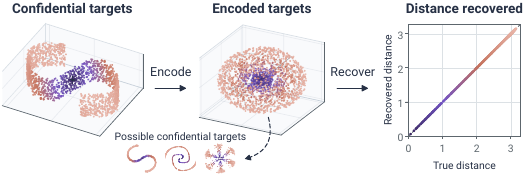}
\caption{Illustrative encoding that preserves each target's distance from the mean (color) while changing its direction.
Different original target shapes can yield the same encoded points, yet these distances remain exactly recoverable.}
\label{fig:property-leakage}
\end{figure}
 
\begin{samepage}
\noindent\textbf{What the radius still reveals.}
\label{cor:property-reconstruction}
Even a high reconstruction error can leave individual properties exposed.
In the illustrative encoding of Figure~\ref{fig:property-leakage}, different original shapes share the same distances from the mean.
The encoded radii reveal these distances exactly, even though the full target values remain ambiguous.
Radius removal discards the magnitude that makes this recovery possible.
\end{samepage}


\subsection{Reconstruction Error after Radius Removal}
\label{sec:leakage}

Section~\ref{sec:ambiguity} showed that the encoded radius still limits privacy: it narrows the indistinguishable targets of each record to those with the same radius.
\name{} therefore removes the radius: the owner divides each rotated encoding by its length and sends only its direction, a unit vector.
Under ideal encoding, the length and direction of a standard Gaussian vector are independent, so the direction reveals nothing about the length.
The indistinguishable target sets can now differ in radius as well as in rotation, so for each record they cover the whole target distribution.
The provider's reconstruction error is therefore at least the full target variance, the error of guessing the mean target for every record.
 
We retain Assumption~\ref{ass:encoding} with measurable encoder and decoder, and let each $\Yout_i$ follow the target distribution.
The provider observes only $\Xin_i$ and
\[
\Ypriv_i=\frac{\encoder(\Yout_i)}{r_i},
\qquad r_i=\|\encoder(\Yout_i)\|_2,
\]
where $r_i>0$ with probability one.
 
Indistinguishable target sets are defined as in Definition~\ref{def:compatible-encodings}, now with the radius-removal form of condition (ii): the direction of $\widetilde{\encoder}(\widetilde{\Yout}_i)$ must equal the observed $\Ypriv_i$.
Since only directions are observed, these sets may change the radius as well as the rotation.
 
\begin{theorem}[Reconstruction bound after radius removal]
\label{thm:radius-removal}
Every attack producing reconstructions $\Yattack_i$ from $\mathcal O$ with finite MSE has an indistinguishable target set $\widetilde{\Yout}$ satisfying
\[
\mathrm{MSE}\bigl(\Yattack,\widetilde{\Yout}\bigr)
\ge\operatorname{Var}(\Yout).
\]
\end{theorem}
 
\begin{proof}
We construct indistinguishable target sets that change the radius while preserving its distribution and the observed unit directions.
Let $F_R$ be the cumulative distribution function of the Gaussian radius $R=\|\encoder(\Yout)\|_2$.
$F_R(R)$ is uniform on $(0,1)$ and independent of direction.
Adding a fixed shift $\delta\in[0,1)$ modulo one preserves this uniform distribution.
Using the same shift and rotation for all records, define
\[
\begin{aligned}
\widetilde r_i&=F_R^{-1}\bigl((F_R(r_i)+\delta)\bmod 1\bigr),\\
\widetilde{\Yout}_i&=\decoder\bigl(\widetilde r_iQ^\top\Ypriv_i\bigr).
\end{aligned}
\]
Null endpoint cases can be assigned any positive radius.
The targets $\widetilde{\Yout}_i$ depend on the chosen pair $(Q,\delta)$.
For each fixed pair, the shifted radius and rotated direction keep their independent Gaussian radius and direction distributions.
Their product is therefore standard Gaussian, and decoding preserves the target distribution.
With $\widetilde{\encoder}(y)=Q\encoder(y)$ and $\widetilde{\decoder}(z)=\decoder(Q^\top z)$, exact inversion gives $\widetilde{\encoder}(\widetilde{\Yout}_i)=\widetilde r_i\Ypriv_i$ with probability one, whose direction is $\Ypriv_i$, so the set is indistinguishable.
 
Fix $\mathcal O$ and the reconstructions $\Yattack_i$, and draw $Q$ and $\delta$ uniformly from $O(k)$ and $[0,1)$, independently of each other, the data, and the attack.
Then $\widetilde r_i$ follows the radius distribution and $Q^\top\Ypriv_i$ is uniform on the unit sphere, independently of $\widetilde r_i$.
Hence $\widetilde{\Yout}_i$ follows the target distribution, and
\[
\begin{aligned}
&\mathbb E\bigl[\|\Yattack_i-\widetilde{\Yout}_i\|_2^2
  \mid\mathcal O,\Yattack_i,r_i\bigr]\\
&\quad=\operatorname{Var}(\Yout)
  +\|\Yattack_i-\mathbb E[\Yout]\|_2^2.
\end{aligned}
\]
The second term is nonnegative.
Averaging over records, data, and attack randomness gives a lower bound of $\operatorname{Var}(\Yout)$ on the error averaged over the random rotation and shift.
At least one fixed pair therefore attains an error at least this large.
\end{proof}
 
The bound equals the error of always guessing the mean target $\mathbb E[\Yout]$.
Compared with Theorem~\ref{thm:reconstruction-bound}, radius removal increases the worst-case reconstruction-error bound by
\[
\operatorname{Var}(\Yout)
-\mathbb E[\operatorname{Var}(\Yout\mid R)]
=\operatorname{Var}\bigl(\mathbb E[\Yout\mid R]\bigr),
\]
i.e., the target variation explained by the radius.
This gain can be zero even when the radius reveals individual properties.
The bounds measure reconstruction of the full target vector and hold in the worst case over indistinguishable target sets, even when the provider knows the target distribution.
 
\subsection{Decoding cost}
\label{cor:radius-cost}
\lc{Need a sentence or two here to introduce this part.}\gs{done}
Radius removal is the only stage that the data owner cannot reverse, so the higher bound comes at a cost in decoding accuracy.
We quantify this cost and show that it becomes small relative to the vectors' length as the latent dimension grows.
To isolate this cost, suppose the released direction is predicted exactly and the owner restores a constant radius before decoding.
For the Gaussian vector $\mathbf w$ before radius removal, with $R=\|\mathbf w\|_2$ and $\Ypriv=\mathbf w/R$, restoring a constant radius $r_0$ gives
\[
\begin{aligned}
\mathbb E[\|\mathbf w-r_0\Ypriv\|_2^2]
&=\mathbb E[(R-r_0)^2]\\
&=\operatorname{Var}(R)+(r_0-\mathbb E[R])^2.
\end{aligned}
\]
The mean radius $r_0=\mathbb E[R]$ therefore minimizes latent reconstruction MSE, leaving error $\operatorname{Var}(R)$.
As the latent dimension $k$ grows, the mean radius grows approximately as $\sqrt{k}$, while its variance approaches $1/2$.
The radii therefore occupy an increasingly narrow band relative to their mean, making a single restored radius a more accurate approximation relative to the vectors' lengths.
Since $\mathbb E[\|\mathbf w\|_2^2]=k$, the relative latent error is $\operatorname{Var}(R)/k$: the error from restoring the mean radius stays bounded while the vectors' expected squared length grows with dimension.
This relative error is about $6.04\%$ at $k=8$ and approaches $1/(2k)$.

\begin{figure*}[!t]
\centering
\includegraphics[width=\textwidth]{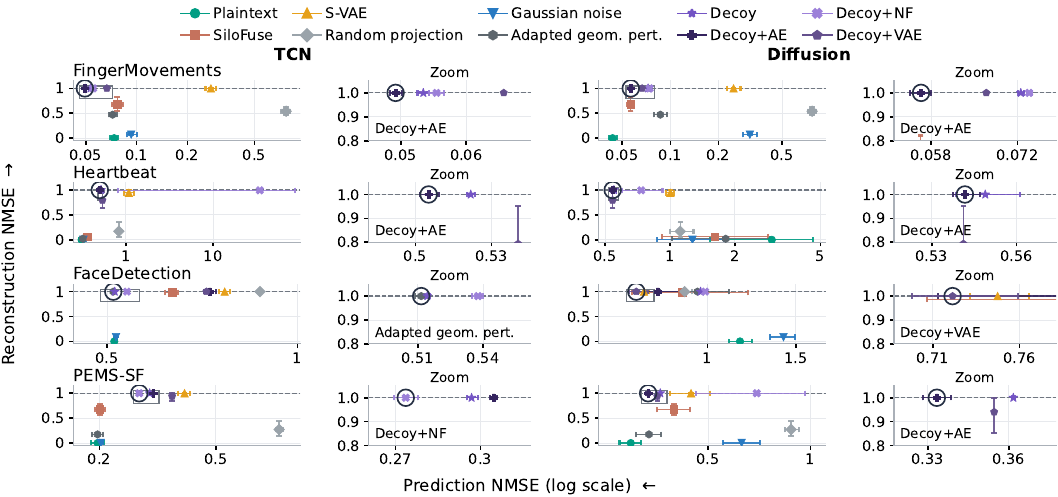}
\caption{Privacy--utility comparison: the preferred corner is top-left.
At a reconstruction NMSE of 1 (dashed line), no tested attack beats the midpoint guess.
The prediction axis is logarithmic.
Each zoom panel enlarges its boxed region; rings mark the largest product $PU$.
All zooms use reconstruction NMSE limits of 0.8--1.05.
Points show means over ten seeds; bars show 95\% bootstrap confidence intervals.}
\label{fig:privacy-utility}
\end{figure*}

\section{Evaluation}
\label{sec:evaluation}

In this section, we evaluate \name{} from several aspects.
First, we assess whether \name{} preserves the accuracy of decoded predictions while limiting the provider's reconstruction of confidential targets.
Second, we measure how much reconstruction improves when the provider obtains confidential targets paired with their encoded targets.
Third, we ablate the encoding stages to identify their contributions to privacy and prediction accuracy.
We also examine how closely the encoded targets follow the uniform distribution on the sphere.

\begin{figure}[t]
\centering
\includegraphics[width=\columnwidth]{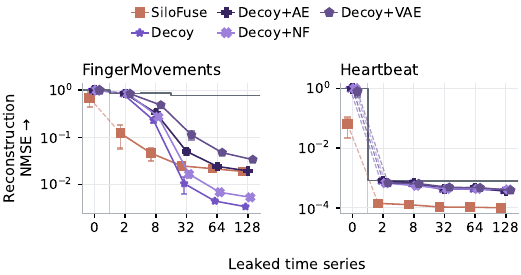}
\caption{Reconstruction NMSE with leaked pairs.
Zero-pair points use the main attacks (ten seeds); positive budgets use paired attacks (three seeds).
Dashed connectors distinguish these settings.
The solid grey step shows the midpoint guess at zero pairs and the leaked-target mean afterward.
Budget positions are equally spaced; small horizontal offsets separate methods at the same budget.
The vertical axis is logarithmic; error bars show sample SD.}
\label{fig:leaked-pairs}
\end{figure}

\begin{figure*}[t]
\centering
\includegraphics[width=\textwidth]{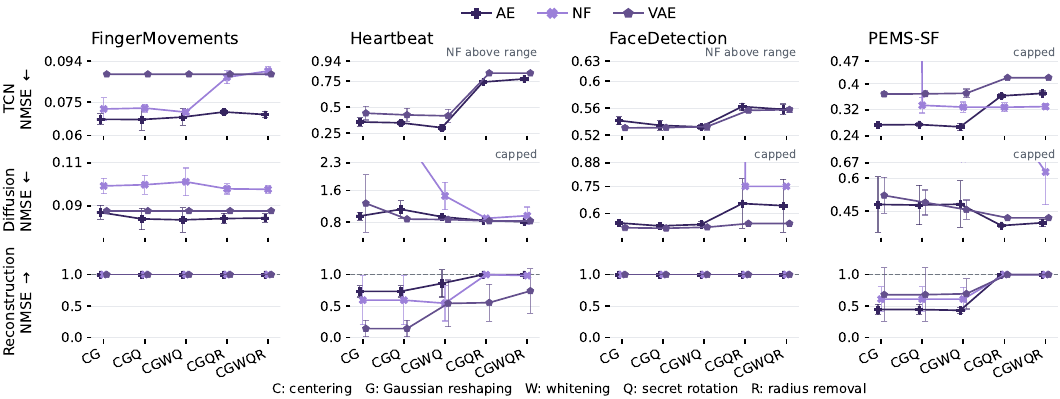}
\caption{Stage ablation for \name{}+AE, \name{}+NF, and \name{}+VAE.
Lines connect mean validation scores over three seeds; error bars show one sample standard deviation.
Prediction axes are linear and capped to make AE and VAE differences visible, so some NF curves lie outside the displayed range.
Error bars are shown only for visible points and stop at the plot boundaries.
A reconstruction NMSE of 1 (dashed line) matches the midpoint guess.}
\label{fig:component-ablation}
\end{figure*}

\subsection{Experimental Setup}
\label{sec:experimental-setup}

\begin{table}[t]
\centering
\caption{Datasets used for channel prediction.}
\label{tab:datasets}
\small
\begin{tabular}{lrr}
\toprule
Dataset & Time series & Channels \\
\midrule
FaceDetection & 9,414 & 144 \\
FingerMovements & 416 & 28 \\
Heartbeat & 409 & 61 \\
PEMS-SF & 440 & 963 \\
\bottomrule
\end{tabular}
\end{table}

\noindent\textbf{Datasets.}
We use four datasets from the UEA multivariate time-series archive~\cite{ueaarchive}: FaceDetection, FingerMovements, Heartbeat, and PEMS-SF (more details in Table~\ref{tab:datasets}).
After discarding the classification labels, we use the first channel as the confidential target and the remaining channels as the shared inputs.
Each time series is then segmented into overlapping 32-step windows using a sliding window with stride one~\cite{rasul2021timegrad}.


\noindent\textbf{Defenses.}
We compare \name{} with learned and parametric defenses.
The learned defenses are the autoencoder of SiloFuse~\cite{silofuse} and S-VAE~\cite{svae}.
The parametric defenses are secret random projection~\cite{randproj}, geometric perturbation adapted with a randomly sampled rotation~\cite{geoperturb}, and additive Gaussian noise~\cite{agrawal2000privacy}.
Training on plaintext targets gives the reference accuracy without encoding.

To assess how the Gaussian reshaping stage affects privacy and prediction, we evaluate four \name{} variants.
Statistical \name{} implements this stage with quantile reshaping followed by whitening.
The three learned variants instead use an autoencoder (\name{}+AE~\cite{hinton2006autoencoder,kolouri2018sliced}), a NICE normalizing flow (\name{}+NF~\cite{nice,nsf}), and a Gaussian variational autoencoder (\name{}+VAE~\cite{kingma2013vae}), and omit whitening.
All four variants retain centering, secret rotation, and radius removal.
\name{}+VAE passes its posterior mean to the later stages, so the encoded targets contain no sampling noise.

\noindent\textbf{Predictors.}
To cover both regression and generation, we use a temporal convolutional network (TCN)~\cite{bai2018tcn} and a conditional diffusion model~\cite{ho2020ddpm,tashiro2021csdi,shankar2025wavestitch}.
Both predictors learn to map shared inputs to encoded targets.
At inference, the diffusion model averages multiple generated samples into a single prediction.

\noindent\textbf{Metrics.}
We report both utility and privacy as a normalized mean squared error (NMSE): the mean squared error (MSE) of a method divided by the MSE of a simple reference guess.
A score of $1$ therefore means the method does no better than this reference.
For utility, we compare the data owner's decoded predictions with the confidential targets.
The reference guess ignores the shared inputs and predicts the test-set mean of each target coordinate.
Lower prediction NMSE indicates better prediction, and $0$ means exact prediction.
For privacy, we compare the provider's reconstructions with the confidential targets and keep the most accurate tested attack.
The reference is the midpoint guess, which uses only the data domain.
Since the midpoint guess is itself counted as an attack, reconstruction NMSE is at most $1$. Lower scores indicate more accurate reconstruction, and thus more leakage.

\noindent\textbf{Evaluation protocol.}
To keep overlapping windows within the same set, we split entire time-series before extracting windows: 20\% of the time-series are reserved for testing and 20\% of the remainder for validation.
All defenses share this split, and each channel is standardized using the mean and standard deviation of the fitting data only.
Hyperparameters of the parametric defenses are selected on the validation sets of FingerMovements and Heartbeat by mean prediction NMSE across both predictors.
Each defense is then refit on the combined training and validation data and evaluated on the test set.
Within each seed, utility and privacy are evaluated on the same sampled encoded targets.
We report means over ten seeds for the main comparison and over three seeds for the leaked-pair and stage-ablation experiments.

For reconstruction, the data domain of each dataset spans the minimum and maximum target values across all splits, rounded slightly outward, which gives the provider tight bounds for every defense.
Against Gaussian noise, the provider also receives the training normalization statistics.

\noindent\textbf{Implementation.}
Our implementation uses Python 3.12 and PyTorch 2.7.0.
All experiments run on a machine with an Intel Core Ultra 7 265K CPU, 64 GB of memory, and an NVIDIA RTX PRO 6000 Blackwell GPU with 96 GB of memory.

\begin{figure*}[t]
\centering
\includegraphics[width=\textwidth]{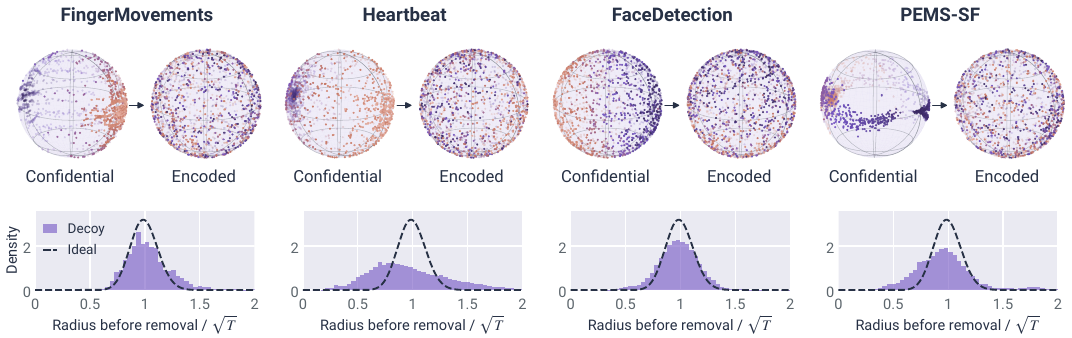}
\caption{Statistical \name{} on the test time series ($T=32$).
Top: directions of confidential ($\Yout$) and encoded ($\Ypriv$) targets on the unit sphere, viewed along each sphere's three directions of largest spread; color orders the records by their coordinate mean (their level), from \textcolor{violet}{violet} (lowest) to \textcolor{orange}{orange} (highest).
Bottom: radius before radius removal; the dashed curve shows standard Gaussian vectors.
\ab{just wondering, doesn't it seem difficult to predict the encoded space?}\gs{it's the privacy-utility tradeoff, in theory the fixed map should still allow training from something} \ab{also, the perhaps more interesting case is DECOY+ neural network encoders}\gs{left for rebuttal}
}
\label{fig:sphere}
\end{figure*}

\subsection{Privacy--Utility Trade-off}
\label{sec:utility}
\label{sec:privacy}


As shown in Figure~\ref{fig:privacy-utility}, \name{} offers the best privacy--utility trade-off in most settings.
We summarize the trade-off with the Nash product $P \times U$ (ringed in each panel), where $P$ is the mean reconstruction NMSE and $U$ is one minus the mean prediction NMSE, so the product is high only when a defense performs well on both objectives.
\name{} variants attain the largest product in seven of the eight dataset--predictor settings, and \name{}+AE leads in five.
Statistical \name{} and \name{}+AE reach a reconstruction NMSE of $1$ on all four datasets, so no tested attack improves on the midpoint guess.


The comparison with SiloFuse is instructive, since both defenses encode targets with an autoencoder.
On FingerMovements with diffusion, \name{}+AE matches the prediction NMSE of SiloFuse while raising reconstruction NMSE from $0.68$ to $1$.
SiloFuse shares its latent representations directly, so their dominant variance direction still tracks the target level.
\name{}+AE instead regularizes the latent space toward a standard Gaussian and removes the radius, which hides this cue while preserving the information needed for prediction.
On FingerMovements and Heartbeat, \name{}+AE also achieves the lowest prediction NMSE among all defenses that reach a reconstruction NMSE of $1$, with either predictor.


Despite these gains, two cases deviate from this pattern.
On FaceDetection with TCN, adapted geometric perturbation narrowly leads statistical \name{}, but both reach a reconstruction NMSE of $1$ and their prediction confidence intervals overlap.
\name{}+VAE is less consistent, with reconstruction NMSE falling to $0.79$ on Heartbeat.
Secret rotation hides the coordinate axes only when the reshaped targets are close to Gaussian, so an imperfect fit by the VAE may leave directional structure that the PCA attack can exploit.


Utility also depends on how the data owner's decoder handles prediction errors.
The provider trains its predictor without access to this decoder and therefore cannot optimize accuracy on the confidential targets directly.
A normalizing flow inverts its outputs exactly, but its inverse can amplify small errors in the predicted encoded targets.
This may explain the unstable TCN predictions of \name{}+NF on Heartbeat, where the mean prediction NMSE reaches $34.1$ with a standard deviation of $83.0$ across seeds.
\name{}+AE does not show this instability, which suggests that exact invertibility is a limited guide to utility: the decoder must also map approximate predictions back to useful targets.

\subsection{Reconstruction from Leaked Pairs}
\label{sec:paired-privacy}



The previous results assume that the provider has no original targets.
Figure~\ref{fig:leaked-pairs} relaxes this assumption and shows how reconstruction changes when the provider obtains confidential targets paired with their encoded targets.
On FingerMovements and Heartbeat, we leak the pairs of every window from $2$ to $128$ training time series.
The provider uses them to train ridge regression and multilayer perceptron decoders, taking the encoded targets alone or together with the shared inputs, and applies these decoders to held-out time series.
Leaked targets also reveal typical target values, so we include their average, the leaked-target mean, as a constant reference.
Comparing against this reference separates what the provider gains from knowing typical values and what it gains from learning to invert the encoding.


On FingerMovements, \name{} resists reconstruction far better than SiloFuse when few time series leak.
With two leaked time series, all \name{} variants stay close to the leaked-target mean at a reconstruction NMSE of about $0.85$, whereas SiloFuse falls to $0.12$.
This gap reflects how directly each encoding exposes the targets.
SiloFuse's latent representations track the target level along a dominant direction, which a decoder can learn from a handful of pairs.
In \name{}, the encoding stages remove this structure, so inverting the encoding requires many more examples.
However, as more time series leak, this advantage disappears, and at $128$ leaked time series \name{}+AE and SiloFuse both reach a reconstruction NMSE of about $0.02$.
In fact, this convergence is expected.
\name{} applies the same fitted maps and secret rotation to every record, so its encoding is a fixed function that the data owner inverts during decoding.
Given enough pairs, the provider can learn this inverse without the encoding parameters.


Heartbeat shows a different pattern, where most of the reduction in reconstruction error comes from the leaked targets instead of the encodings.
Two leaked time series already make the leaked-target mean over a thousand times more accurate than the midpoint guess.
The targets thus occupy a small part of the data domain, so knowing typical values is enough to bring the provider close to them.
\name{} variants remain near the leaked-target mean across all budgets, meaning that the encoded targets reveal little beyond these typical values.
In contrast, SiloFuse lets attacks reduce the error by a further factor of six to eight.


These results mark the boundary of \name{}'s threat model, which assumes that the provider observes no pairs of targets and their encodings.
Within this assumption, no tested attack improves on the midpoint guess, and when only a few pairs leak, \name{} still limits reconstruction to what typical target values reveal.

\subsection{Stage Ablations}
\label{sec:ablations}


To identify which stages provide \name{}'s protection, we ablate whitening, secret rotation, and radius removal in the three learned variants.
Figure~\ref{fig:component-ablation} compares five stage combinations, all of which include centering and the Gaussian reshaping performed by the learned encoder.
Whitening, which the learned variants omit in the main comparison, is added back here to test whether it brings further benefit.
Within each dataset, variant, and seed, all combinations reuse the same fitted maps and rotation key, so any difference between them comes from the varied stages.
We evaluate all combinations on validation data.


Radius removal is the stage that most improves resistance to reconstruction.
For \name{}+AE, adding it after rotation raises reconstruction NMSE from $0.73$ to $1$ on Heartbeat and from $0.44$ to $1$ on PEMS-SF, and \name{}+NF and \name{}+VAE likewise reach $1$ on PEMS-SF.
This result matches the analysis in Section~\ref{sec:theory}.
Rotation preserves the radius of each encoded target, and this radius narrows the set of targets consistent with each encoding.
Once the radius is removed, only the direction remains, which under Gaussian reshaping carries no information about the radius.
On FingerMovements and FaceDetection, all five combinations already reach a reconstruction NMSE of $1$, since the tested attacks fail after centering and Gaussian reshaping.


Secret rotation and whitening show no measurable effect on reconstruction.
For rotation, this is expected.
The range-only PCA attack follows the principal axes wherever they point, so rotating the encoded targets changes neither the axes it finds nor its error.
Rotation instead hides the orientation of the fitted maps, which a provider that knows the data domain could otherwise approximate, as formalized in Proposition~\ref{thm:ambiguity}.
Radius removal complements this role by discarding the magnitude that rotation preserves.
Whitening brings no further benefit because the learned encoders already map all coordinates jointly toward a standard Gaussian.


Turning to utility, although radius removal is the only stage that discards information, it does not consistently reduce prediction accuracy.
For \name{}+AE on Heartbeat and PEMS-SF, it improves diffusion accuracy along with reconstruction resistance, although it increases TCN error.
The discarded information is small, because the radii of Gaussian vectors concentrate around their mean as the dimension grows, so restoring the mean radius during decoding introduces little error (Section~\ref{sec:theory}).
The opposite effects on TCN and diffusion show that the net impact on utility depends on the predictor.

\subsection{Geometry of Encoded Targets}
\label{sec:geometry}


Our analysis in Section~\ref{sec:theory} assumes ideal encoding, where Gaussian reshaping and whitening turn confidential targets into standard Gaussian vectors, and secret rotation and radius removal leave them uniformly distributed on the unit sphere.
Real targets satisfy this assumption only approximately, so Figure~\ref{fig:sphere} examines how closely statistical \name{} approaches it on the test time series.
Each set of vectors is viewed along its three directions of largest spread, where departures from uniformity are largest.


Before encoding, confidential targets occupy small regions of the sphere, and their position follows their coordinate mean.
On FingerMovements and Heartbeat, records with low and high coordinate means gather near opposite poles, and on PEMS-SF they lie along a narrow band.
Even on FaceDetection, where the targets spread more widely, records remain ordered by their coordinate mean along the main axis.
This structure arises because differences in target level dominate the variation across records, which is exactly the cue that the range-only PCA attack exploits.
After encoding, the targets cover the whole sphere on all four datasets, and records with similar coordinate means are scattered across it.
Gaussian reshaping and whitening remove the dominant direction, so the direction of an encoded target no longer indicates its level, and the PCA attack does not improve on the midpoint guess.

The radii before radius removal show where the encoding still departs from standard Gaussian vectors.
Their mean lies between $0.98\sqrt{T}$ and $1.04\sqrt{T}$, as expected, but their standard deviation relative to the mean ranges from $0.19$ on FaceDetection to $0.45$ on Heartbeat, compared with $0.125$ for standard Gaussian vectors.
This excess spread likely arises because reshaping and whitening make the coordinates marginally Gaussian and uncorrelated, but not independent, so coordinates that move together enlarge or shrink the radius jointly.

Radius removal discards this spread, and the remaining directions appear close to uniform.
Real targets thus leave more information in the radius than the ideal analysis predicts, which is consistent with the stage ablation, where radius removal most improves resistance to reconstruction.


\section{Conclusion}
\label{sec:conclusion}

Existing encodings can expose confidential targets through the same patterns they preserve for learning.
We develop two attacks that exploit these patterns and introduce \name{}, which reshapes targets, applies a secret rotation, and removes radius information.
Under ideal Gaussian encoding, we establish worst-case reconstruction-error lower bounds even when the provider knows the target distribution.
Across four time-series datasets, attacks on existing defenses achieve up to 95\% lower reconstruction error on average than guessing the midpoint of the known target range.
For statistical \name{} and \name{}+AE, no tested range-only attack improves on this baseline.
These encodings preserve useful patterns for prediction while limiting the reconstruction accuracy of the tested attacks.

\appendix
\begin{ethics}
The provider receives encoded targets to train a predictor, but it may also use them to reconstruct the confidential targets.
We studied this risk using four public time-series datasets.
A provider could use the published attacks on encoded targets it has already received.
We also used these attacks to evaluate \name{}.
On all four datasets, none of the tested attacks improved on the naive guess when the provider knew only the target range.
We consider releasing the attacks alongside \name{} justified because data owners can use them to check for this exposure before sharing their targets and to evaluate a defense against it.
The study uses only public datasets and involves no human participants or personal data, so we did not seek review from an ethics board.
\end{ethics}

\begin{openscience}
The code for our forecasting experiments and reconstruction attacks is available at \url{https://anonymous.4open.science/r/decoy-BA4D/}.
The repository includes scripts to download the four public datasets and rerun the main comparison, stage ablations, and reconstruction experiments with leaked pairs.
\end{openscience}

\begin{ai}
We utilized GPT-6-Astra (OpenAI) and Claude Opus 5.5 (Anthropic) to assist with the implementation of experimental methods, development of data visualization scripts, and orchestration and monitoring of long-running experiments.
We did not use AI tools for data curation or generation, drafting parts of the manuscript or formulating hypotheses.
All decisions, analysis, and conclusions were made or fully verified by humans.
\end{ai}

\bibliographystyle{ACM-Reference-Format}
\bibliography{references}

\section{Reconstruction Error of the Range-Only PCA Attack}
\label{sec:pca-bound}

The range-only PCA attack estimates the level of each confidential target from its encoded target.
Section~\ref{sec:attacks} bounds the error when the PCA scores are close to the target levels after a shift and a change of scale.
Here we give a condition on the encoded targets under which the scores approximate those levels, and show how the error in estimating a level affects reconstruction of the full target.

Let $h_i=\frac1T\sum_{t=1}^T Y_{i,t}$ be the level of target $\Yout_i$, and let $\bar h$ be the average level across records.
Define $\mathbf u=(h_1-\bar h,\ldots,h_n-\bar h)^\top$ and $\sigma=\|\mathbf u\|_2/\sqrt n$, the standard deviation of the target levels.
Let $M\in\mathbb R^{n\times k}$ contain the centered encoded targets as rows.
For $\mathbf u\ne0$, fitting each column of $M$ as a linear function of $\mathbf u$ gives
\[
F=\frac{\mathbf u(\mathbf u^\top M)}{\mathbf u^\top\mathbf u}.
\]
\begin{assumption}[Linear structure in encoded targets]
\label{ass:pca-linear-structure} 
The target levels are not all equal, $F\ne0$, and
\[
\rho=\frac{\|M-F\|}{\|F\|}<1,
\]
where $\|\cdot\|$ denotes the largest singular value.
\end{assumption}
Under this assumption, differences in target levels account for more variation along the fitted direction than the residual accounts for along any single direction.

To separate PCA error from range calibration, let $\mathbf v_1$ be a unit first principal direction of $M$.
We choose its sign so that $F\mathbf v_1=\alpha\mathbf u$ with $\alpha>0$.
The adjusted scores $s_i=\bar h+(M\mathbf v_1)_i/\alpha$ express the PCA scores in the same units as the target levels.
Let $g$ be the linear rescaling that maps the smallest and largest adjusted scores to $a-\Delta$ and $b+\Delta$, respectively.
We measure how far this calibration moves the scores by
\[
C=\left(\frac1n\sum_{i=1}^n (g(s_i)-s_i)^2\right)^{1/2}.
\]
The true levels and the adjusted scores enter only the analysis.
The provider applies min--max scaling directly to the PCA scores.

\begin{theorem}[Range-only PCA reconstruction error]
\label{thm:pca-bound}
Suppose Assumption~\ref{ass:pca-linear-structure} holds, and let $\widehat h_i$ denote the attack's estimates using $[a-\Delta,b+\Delta]$.
After possibly reflecting all estimates as $\widehat h_i\mapsto a+b-\widehat h_i$,
\begin{equation}
\label{eq:pca-property-bound}
\frac1n\sum_i\bigl(\widehat h_i-h_i\bigr)^2
\le\left(C+\frac{\sigma\rho}{\sqrt{1-\rho^2}}\right)^2.
\end{equation}
For the same reflection, repeating each estimate across all $T$ coordinates gives reconstructions $\Yattack_i$ with
\begin{equation}
\label{eq:pca-value-bound}
\begin{aligned}
\mathrm{MSE}(\Yattack,\Yout)
&\le\frac1n\sum_i\sum_{t=1}^T\bigl(Y_{i,t}-h_i\bigr)^2\\
&\quad+T\left(C+\frac{\sigma\rho}{\sqrt{1-\rho^2}}\right)^2.
\end{aligned}
\end{equation}
Here, MSE averages squared Euclidean error over the $n$ records.
\end{theorem}

\begin{proof}
Least squares gives $\mathbf u^\top(M-F)=0$.
For every unit vector $\mathbf v$, the fitted scores and residual scores are therefore orthogonal, giving
\[
\|M\mathbf v\|_2^2=\|F\mathbf v\|_2^2+\|(M-F)\mathbf v\|_2^2.
\]
In particular, $\|M\|\ge\|F\|$.
Since PCA maximizes $\|M\mathbf v\|_2$ over unit vectors,
\[
\|F\mathbf v_1\|_2^2\ge(1-\rho^2)\|F\|^2>0.
\]
Thus $F\mathbf v_1$ is a nonzero multiple of $\mathbf u$, and we may choose its sign as above.
Subtracting $h_i$ from $s_i$ leaves the $i$th entry of $(M-F)\mathbf v_1/\alpha$, so
\[
\sqrt{\frac1n\sum_i(s_i-h_i)^2}
\le\frac{\sigma\|M-F\|}{\|F\mathbf v_1\|_2}
\le\frac{\sigma\rho}{\sqrt{1-\rho^2}}.
\]
After centering, the adjusted scores equal the nonzero vector $\mathbf u$ plus an orthogonal residual.
Their range is therefore nonzero.
Min--max scaling cancels the affine adjustment up to reflection.
The triangle inequality then adds $C$ to the root mean squared error.
Finally, $\sum_t(Y_{i,t}-h_i)=0$, so for each record
\[
\|\Yattack_i-\Yout_i\|_2^2
=\sum_{t=1}^T(Y_{i,t}-h_i)^2+T(\widehat h_i-h_i)^2.
\]
Averaging over records gives~\eqref{eq:pca-value-bound}.
\end{proof}

Even if the scores recover the relative levels exactly, mapping them to the known range can introduce error when the observed levels cover only part of that range.
This error is captured by $C$.
Widening the range from $[a,b]$ to $[a-\Delta,b+\Delta]$ moves each calibrated estimate by at most $\Delta$, so it increases $C$ by at most $\Delta$.
The second bound shows why recovering the level can reveal individual values when the coordinates of a target vary little around their mean.

A shared orthogonal rotation of the encoded targets preserves $MM^\top$ and hence the PCA scores up to sign when the leading singular value is simple.
It also rotates the fit and the residual together, preserving $\rho$.
The sign ambiguity concerns the whole set of records, so the same reflection must be used for every estimate.

\section{Reconstruction Error of the Predictor-Denoising Attack}
\label{sec:denoising-bound}

Under additive noise, the provider can use its prediction of the encoded target directly as a reconstruction.
We consider a predictor trained on noisy targets and ask when it improves on guessing the midpoint for a record from an independent recording.

Let $f$ be the fitted predictor and $m(\Xin)=\mathbb E[\Yout\mid\Xin]$ the conditional mean of the confidential target.
As in Section~\ref{sec:predictor-denoising}, the encoded target is $\Ypriv=\Yout+\eta$, with $\mathbb E[\eta\mid\Xin]=0$.
It follows that $m(\Xin)$ also minimizes the expected squared error for predicting $\Ypriv$.
Write $\mathbf c$ for the vector whose $T$ coordinates all equal the midpoint $(a+b)/2$.

\begin{proposition}[Improvement over the midpoint guess]
For a fixed fitted predictor $f$, let $(\Xin,\Yout)$ be drawn independently of its training data from the same distribution.
Assume the targets, noise, and predictions have finite second moments.
The reduction in expected reconstruction error is
\[
\begin{aligned}
&\mathbb E\|\Yout-\mathbf c\|_2^2-\mathbb E\|\Yout-f(\Xin)\|_2^2\\
&\quad=\mathbb E\|m(\Xin)-\mathbf c\|_2^2
       -\mathbb E\|f(\Xin)-m(\Xin)\|_2^2.
\end{aligned}
\]
\end{proposition}

\begin{proof}
Given $\Xin$, the residual $\Yout-m(\Xin)$ has mean zero.
Its expected inner product with $m(\Xin)-f(\Xin)$ is therefore zero.
Expanding the squared reconstruction error gives
\[
\mathbb E\|\Yout-f(\Xin)\|_2^2
=\mathbb E\|\Yout-m(\Xin)\|_2^2
 +\mathbb E\|m(\Xin)-f(\Xin)\|_2^2.
\]
The same expansion holds for the constant guess $\mathbf c$.
Subtracting cancels the residual error.
\end{proof}

The first term in the proposition is the improvement the provider would obtain if it knew the conditional mean.
The fitted predictor beats the midpoint exactly when its error in estimating this mean is smaller than that improvement.
Added noise can affect how accurately the provider learns the mean from finitely many training records.
For reconstructions on those same records, the fitted predictor also depends on their encoded targets, so the independent-record calculation does not give their reconstruction error.

\end{document}